\documentclass{article}
\usepackage{spconf,amsmath,graphicx}
\usepackage{amssymb}
\usepackage{xspace}
\usepackage{booktabs}
\usepackage{cleveref}
\usepackage{multirow}

\usepackage{acronym}

\acrodef{LM}{language model}
\acrodef{uLM}{unit Language Model}
\acrodef{SOTA}{State-Of-The-Art}
\acrodef{GSLM}{Generative Spoken Language Modeling}
\acrodef{TTS}{Text-to-Speech}
\acrodef{WER}{Word-Error-Rate}
\acrodef{ASR}{Automatic Speech Recognition}
\acrodef{SSL}{Self-Supervised Learning}

\usepackage{subcaption}

\newcommand{\method}{\textsc{LAST}\xspace}
\newcommand{\slm}{SpeechLM\xspace}
\newcommand{\slms}{SpeechLMs\xspace}

\usepackage{etoolbox}
\newtoggle{release}

\togglefalse{release}
\toggletrue{release}

\iftoggle{release}{
\newcommand{\adios}[1]{}
\newcommand{\AT}[1]{}
\newcommand{\ms}[1]{}
\newcommand{\rid}[1]{}
\newcommand{\skp}[1]{} 
}
{
\newcommand{\adios}[1]{\textcolor{red}{YA: #1}}
\newcommand{\AT}[1]{\textcolor{magenta}{AT: #1}}
\newcommand{\ms}[1]{\textcolor{red}{M: #1}}
\newcommand{\rid}[1]{\textcolor{red}{$<$#1$>$}}
\newcommand{\skp}[1]{\textcolor{orange}{#1}}
}
\newcommand{\newpara}[1]{\vspace{0.3cm}\noindent \textbf{#1}}

\title{\method: \underline{L}anguage Model \underline{A}ware \underline{S}peech \underline{T}okenization}
\name{Arnon Turetzky and Yossi Adi}
\address{
    School of Computer Science and Engineering\\
    Hebrew University of Jerusalem\\
    \texttt{\{arnon.turetkzy\}@mail.huji.ac.il}
    }

\begin{document}

\maketitle

\begin{abstract}
Speech tokenization serves as the foundation of speech \ac{LM}, enabling them to perform various tasks such as spoken language modeling, text-to-speech, speech-to-text, etc. Most speech tokenizers are trained independently of the \ac{LM} training process, relying on separate acoustic models and quantization methods. Following such an approach may create a mismatch between the tokenization process and its usage afterward. In this study, we propose a novel approach to training a speech tokenizer by leveraging objectives from pre-trained \emph{textual} \ac{LM}s. We advocate for the integration of this objective into the process of learning discrete speech representations. Our aim is to transform features from a pre-trained speech model into a new feature space that enables better clustering for speech \ac{LM}s. We empirically investigate the impact of various model design choices, including speech vocabulary size and text \ac{LM} size. Our results demonstrate the proposed tokenization method outperforms the evaluated baselines considering both spoken language modeling and speech-to-text. More importantly, unlike prior work, the proposed method allows the utilization of a single pre-trained \ac{LM} for processing both speech and text inputs, setting it apart from conventional tokenization approaches.
\end{abstract}

\begin{keywords}
Speech Tokenization, Speech Language Models
\end{keywords}

\section{Introduction}
\label{sec:intro}

The development of Speech Language Models (\slms) was recently raised as a new research direction within the spoken language processing community~\cite{on_generative, kharitonov2022textless, park2023generative, kharitonov2021text, borsos2022audiolm, qian2022contentvec, algayres2023generative, wu2023improving}. \slms are usually composed of two or three main components: (i) speech tokenizer which converts raw speech signals into discrete tokens; (ii) \ac{uLM}, operating over this discrete representation to learn the underlying distribution of speech utterances; and (iii) in the generative setup, a unit-based vocoder which converts speech tokens into a waveform signal~\cite{on_generative, polyak2021speech}. The ability to operate directly over raw speech recordings without accessing any textual supervision holds great potential: (i) This can be beneficial for languages that do not have large textual resources or standardized orthography (Swiss German, dialectal Arabic, Igbo, etc.); (ii) It may also be useful for ``high-resource'' languages, in which the oral and written forms often mismatch; and (iii) It can also model non-verbal cues such as laughter, coughing, etc. Recent studies in the field have shown that following this modeling paradigm can be beneficial for spoken dialogue modeling~\cite{nguyen2022generative}, speaking style conversion and speech emotion conversion~\cite{kreuk2021textless, maimon2022speaking}, direct speech-to-speech translation~\cite{lee2021direct, lee-etal-2022-textless, popuri2022enhanced, peng2024mslm}, silent video-to-speech generation~\cite{hsu2022revise}, etc. In this work, we study the speech tokenization part. 

The most common approach nowadays to speech tokenization is applying the k-means algorithm over representations obtained from a pre-trained \ac{SSL} models, this results in a \emph{semantic speech tokens}~\cite{on_generative, borsos2022audiolm}. While being simple and effective, this approach has several drawbacks. First, it requires a separate training stage on top of the \ac{SSL} model, and can not be done jointly. Second, it is not clear what layer should we use to obtain representations to train the k-means model. Prior work proposed different layers~\cite{on_generative, kharitonov2021text, nguyen2023expresso} and show varying results when different layers are being chosen, even when considering the same model. Lastly, the k-means model is sensitive to the data used to learn the cluster centroids. Hence, it remains an open question of what data should be used to train the clustering model. Prior work found that different dataset splits result in different model performance~\cite{nguyen2023expresso}. Another line of speech and audio tokenization was proposed by \cite{defossez2022high, zeghidour2021soundstream} using a Residual Vector Quantization (RVQ) method under the auto-encoding setup. Such a tokenization method is highly general and can capture any type of audio (not only speech), however, it was found to be challenging to optimize \slms on top of it without conditioning on text or semantic tokens~\cite{borsos2022audiolm}. Such representation is often denoted as \emph{acoustic tokens}. 

\begin{figure*}[t!]
  \centering
  \includegraphics[width=\linewidth]{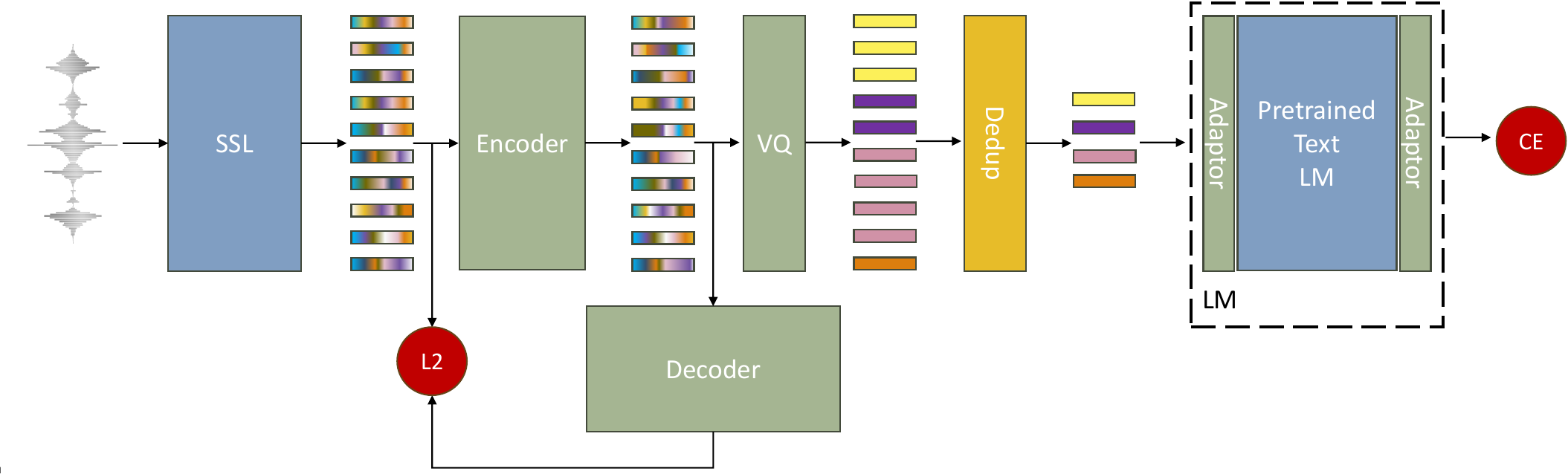}
  \caption{\textbf{A visual description of \method.} We propose to leverage a pre-trained text-LM to construct a speech tokenizer. \method receives gradients from the LM to guide the tokenization process toward better sequence modeling. Pretrained freezed modules are blue and learned modules are green.}
  \label{fig:method}
\end{figure*}

Considering the fact that the above-mentioned tokens will be later on used to train \slms, raises the following question: \emph{can we construct a speech tokenizer which will be guided by a language model?} Moreover, recently~\cite{hassid2024textually} found that performing a warm-initialization of \slms parameters from a pre-trained text \ac{LM}, such as OPT~\cite{zhang2022opt} or LLaMA 2~\cite{touvron2024llama}, results in superior performance to cold-random initialization. 

Equipped with the above findings we propose \method, which stands for \underline{L}anguage Model \underline{A}ware \underline{S}peech \underline{T}okenization. Specifically, we propose to involve a pre-trained text \ac{LM} during the tokenization process to result in speech tokens that will be better suited for sequential modeling. Generally, \method is comprised of three main components: (i) a pre-trained, frozen speech \ac{SSL} model which extracts contextualized speech representations from speech signals. We experimented with HuBERT~\cite{hubert}, however \method is general and can be applied to any \ac{SSL} method; (ii) an adapter-quantization module which converts this contextualized representation into discrete tokens; and (iii) a pre-trained, frozen \ac{LM} which guides the tokenization process towards better sequential modeling. A visual description of the proposed method can be seen on \Cref{fig:method}. Notice, in contrast to~\cite{hassid2024textually} which fine-tunes text \ac{LM} using speech tokens, and as a result removing the learned textual information, the proposed approach keeps the text-LM frozen, hence does not affect performance on text benchmarks. 

We evaluate \method on a set of zero-resource speech modeling tasks~\cite{nguyen2020zero} and found the proposed method is superior to the traditional k-means method across all setups. We also evaluate the ability of the newly proposed speech tokens on the task of \ac{ASR} and found them to produce superior performance than the k-means alternative. We additionally provide an extensive ablation study, which sheds light on the importance of each of the components constructing \method.

\section{Background}
\label{sec:back}

\begin{figure*}[t!]
  \centering
  \includegraphics[width=\linewidth]{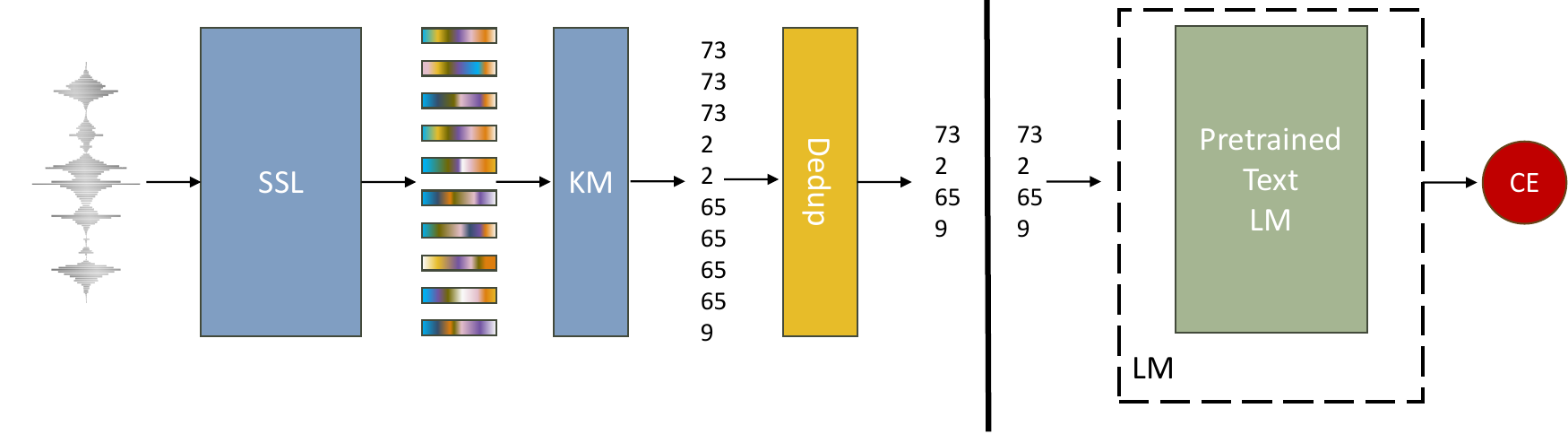}
  \caption{\textbf{The common \emph{SpeechLM} pipeline.} First, discrete representation is extracted from the raw waveform using both a speech encoder and a quantization module (often known as a speech tokenizer). This representation is later used for training a \ac{uLM}. In this study, we focus on the speech tokenization part.}
  \label{fig:gslm}
\end{figure*}

As mentioned before, the general pipeline for constructing \slms is comprised of two main modules: (i) Speech tokenizer and (ii) \ac{uLM} (See Figure~\ref{fig:gslm} for a visual description). In the following paragraphs, we give a background for each of the components including the standard evaluation methods.  

\newpara{Speech Tokenizer} module encodes the raw speech signal into a discrete representation. The common approach is first to encode the speech into a continuous representation and then quantize the representation to achieve a sequence of discrete units~\cite{on_generative, polyak2021speech, popuri2022enhanced, lee2021direct, kharitonov2021text, kreuk2021textless, kharitonov2022textless, nguyen2022generative, borsos2022audiolm, tjandra2019vqvae, tjandra2020transformer}. 

Formally, denote the domain of audio samples by $\mathcal{X} \subset \mathbb{R}$. The representation for a raw signal is therefore a sequence of samples $x = (x_1,\ldots, x_T)$, where  $x_t\in\mathcal{X}$ for all $1\leq t \leq T$. 

Consider an encoder network, $f$, that gets as input the speech utterance and outputs a sequence of spectral representations sampled at a low frequency as follows $f(x) = (v_1, \dots, v_{T'})$. Note that we do not assume anything about the structure of the encoder network $f$. \cite{on_generative}, evaluated several speech encoders, namely, Mel-spectrogram, Contrastive Predictive Coding~\cite{oord2018representation}, wav2vec2~\cite{baevski2020wav2vec}, and HuBERT~\cite{hubert}. 

Since the representations learned by such models are usually continuous, a tokenization or quantization algorithm is applied over the models' outputs to generate discrete units, denoted as $z = (z_1,\ldots,z_{T'})$. Each element $z_i$ in $z$ is a positive integer, $z_i\in\{1,...,K\}$ for $1\le i \le T'$, where $K$ is the number of discrete units. We denote the quantization model with $Q$. The common approach in prior work is to use the k-means algorithm as the quantization method. 

\newpara{Unit Language Model}  is trained on the extracted discrete units, $z$. Such a language model learns a probability distribution of the learned unit sequences, which enables direct modeling of speech data without textual supervision. 

The language model can be used to sequentially model speech utterances, and generate speech conditionally or unconditionally. Moreover, such a modeling framework allows for capturing and modeling prosodic features~\cite{kharitonov2021text}, as well as speaker identity~\cite{borsos2022audiolm}, or even natural dialogues~\cite{nguyen2022generative}. This is in contrast to using textual features, as they do not encode such information. In this work, we do not focus on speech generation but rather demonstrate that the proposed speech tokenizer provides better performance for modeling sequential data via the \ac{uLM}.  

\newpara{Zero-Resource Speech Evaluation.} \cite{nguyen2020zero} proposed a set of zero-shot evaluation tasks specifically targeting speech modeling (i.e., sWUGGY, and sBLIMP). The sWUGGY metric requires detecting the real word from a pair of short utterances such as 'brick' vs. 'blick.' Similarly, sBLIMP requires detecting the syntactically correct sentence from a pair of sentences. In both metrics, detection is done by comparing the probabilities of both sequences. These metrics are desirable under our setup as, unlike perplexity, these allow us to compare models with different tokenizers. We mainly report this method throughout the paper.

\section{Language Aware Speech Tokenizer}
\label{sec:method}
In this section we present \method. We start by describing the speech tokenization approach guided by a frozen LM, followed by the exact implementation details and fine tuning configuration. 

\subsection{Model}
\label{subsec:model}

Given a speech utterance $x$, we first feed it into a frozen and pre-trained speech encoder network $f(x) = v = (v_1, \dots, v_{T'})$, where each $v_i\in\mathbb{R}^d$. Next, $v$ is being fed into a learnable encoder module $E(v) = u = (u_1, \dots, u_{T'})$, where each $u_i\in\mathbb{R}^{d'}$. Then, a quantization process is performed, via a Vector Quantization (VQ) module, $Q$, to quantize each $u_i$. Formally, $Q(u) = (z_1,\dots,z_{T'})$, where each $z_i$ is a positive integer in the range of $\{1, \dots, K\}$ and $K$ is the number of codes in $Q$. 

Next, to guide the quantization process toward sequential modeling, we feed $z$ into a pre-trained textual LM to perform next token prediction. To extend the text LM for speech tokens, we add randomly initialized adaptation layers before and after the LM which we freeze for the entire training, hence keeping the existing text LM capabilities unchanged. We analyze the effect of the adapter size in \Cref{subsec:adaptors}. 

Similarly to the common practice in training \ac{LM}s we minimize the negative log-likelihood loss between the predicted and true probability distributions over the learned tokens. Formally, we minimize the following, 

\begin{equation}
    \mathcal{L}_{LM} = -\sum_{i=1}^{n} \log p_{\theta}(z_i | z_{i-1}, \ldots, z_1),
    \label{eq:lm_loss}
\end{equation}

where we consider $\theta$ as the parameters of both the adaptor and new speech tokens look-up table. We do not update the LM parameters, we only backpropagate through it. 

Notice, that we learn to tokenize speech via next token prediction. Such an optimization process may easily collapse to a single token or a sequence of tokens. To prevent collapse we introduce a reconstruction loss function to stabilize the optimization process. For that, we introduce a decoder module, $D$, which gets as input $u$ and is optimized to reconstruct $v$. Specifically, we minimize the L2 loss between $D(u)$ and $v$. Overall, the objective function of the system is as follows, 

\begin{equation}
    \mathcal{L} = \mathcal{L}_{LM} + \lambda \|D(u) - v\|_2, 
    \label{eq:loss}
\end{equation}

where $\lambda$ is a hyperparameter balancing between both loss functions. A visual description of the proposed method can be seen in \Cref{fig:method}.

\begin{table}[t!]
\centering
\begin{tabular}{lccccc}
\toprule
\textbf{Model} & \textbf{TWIST} & \multicolumn{2}{c}{\textbf{sWUGGY} $\uparrow$} & \textbf{sBLIMP$\uparrow$} \\
 & & Inter & OOV &  \\
 \midrule
k-means\rid{39}&$\times$ & 63.53 & 54.06 & 53.3 \\
k-means\rid{38}& \checkmark & 72.61 & 57.47 & 55.73\\
\midrule
\method\rid{41}&$\times$ & 70.79 & 57.13 & 54.76\\
\method\(^\dag\)\rid{6}& $\times$ & 71.78 & 57.19 & 55.25\\
\method\rid{25,\textbf{42}}& \checkmark & 74.24 & 58.37 & 56.80\\
\bottomrule
\end{tabular}
\caption{
Zero-resource speech metrics for sequence modeling. We present results for both k-means tokenizer and \method with and without finetuning the \ac{LM}. Results are presented using an OPT model with $350$M parameters and a codebook size of 500. TWIST refers to fine-tuning the \slm initialized from text LM parameters. \method with adaptation layers marked with \(\dag\)}
\label{LM-results}
\end{table}

\subsection{Details}
\label{subsec:training}
We set $f$ to be a pre-trained HuBERT-\texttt{base} can be beneficial
for languages that do not have large textual resources or even a widely used standardized
orthography{base} model \cite{hubert}. Similarly to~\cite{hassid2024textually}, representations are obtained from its 9th layer. For $E$, we utilize a transformer encoder followed by a projection layer, where the latent dimension $d'$ is determined by the LM dimension.

We employ VQ \cite{van2017neural, defossez2022high} to discretize the output $u$ where we experimented number of codebooks $\in \{100, 200, 500, 1000\}$. We empirically analyze results for the different number of codebooks in \Cref{subsec:vocab_size}.

For the LM part, we utilize the OPT model \cite{zhang2022opt} with $125$M and $350$M parameter models. These models are initialized from a pre-trained text \ac{LM} following \cite{hassid2024textually}. We employ a separate lookup table with with a new randomly initialized embedding and add two more projection layers to enable speech tokens in addition to text tokens. In order to feed tokens into the model as input, the first layer projects from the embedding dimension to the model dimension. The second layer maps the model output to the number of tokens, allowing for the calculation of next speech token probabilities. In our setup we use only the speech tokens but using both text and speech tokens, only requires an indicator to specify whether to use text or speech. This indicator will determine which group of layers should be used for each token. Similarly to \cite{lakhotia2021generative} we remove sequential repetitions of speech tokens before feeding them into the \ac{LM}. Notice, unlike \cite{hassid2024textually} which fine-tunes the text LM and as a result removes text modeling capabilities, following the proposed approach keeps the original performance over text benchmarks unchanged while extending it to speech modeling ones. We update the \ac{LM} lookup table with the vector quantization codebooks at each training step and feed $z$ as input for the \ac{LM}.

\newpara{Finetune.} Although the proposed method allows the utilization of a single pre-trained \ac{LM} for processing both speech and text inputs, we would also evaluate \method similarly to \cite{hassid2024textually}, i.e., finetuning the LM. We denote the difference between both cases as ``Pretrain'' and ``Finetune'' to avoid confusion. 
\section{Experimental Setup}
\label{sec:setup}

We use different numbers and types of GPUs across our experiments due to diverse computational requirements ranging from single to 4 GPUS out of the Nvidia A5000, A6000, and A40. Due to computational constraints we use grad accumulation, and set a limit for maximum steps. Unless stated otherwise, except for the speech recognition experiments, we set a limit of $200K$ optimization steps with a batch size of $32$. We sampled $10$ seconds from each example and zero padded from right whenever needed. We used AdamW optimizer and linear-warmup-cosine scheduler that increase learning rate from $0$ to $1e-4/2.5e-5$ by $1K$/$10K$ steps then decrease to $3e-5$/$1e-5$ for pretraining and finetuning experiments respectively. Speech recognition experiments were limited to $150K$ steps with batch size $64$, same optimizer and scheduler but with a warmup of $5K$ steps to learning rate $1e-4$ decrease to $1e-5$.

\section{Results}
\label{sec:res}

We report results for the zero-resource speech metrics~\cite{dunbar2017zero, nguyen2020zero} together with speech recognition metrics, comparing the proposed method to the k-means alternative. 

\subsection{Zero-Resource Speech Metrics}
We start by evaluating the proposed method using the zero-resource speech metrics as proposed by~\cite{nguyen2020zero}, namely sWUGGY and sBLIM, and compare it to the commonly used k-means tokenizer. Unless stated otherwise, for a fair comparison we follow the same approach as in \cite{lakhotia2021generative} and report the results for the ``in-vocabulary'' split. For both measures, we compare the geometric mean of the models' sequence probabilities assigned to each utterance within a pair. We report results in two setups. In the first one, we keep the LM frozen and do not update its parameters. In the second setup, we follow TWIST~\cite{hassid2024textually}, and we fine-tune the LM initialized from a text model. For a fair comparison to the k-means baseline, we consider two setups of \method: with and without the adaptation layers. Both \method and the k-means baseline are built on top of the HuBERT-\texttt{base} model. Results are presented in \Cref{LM-results} using with OPT-$350$ model and codebook size of 500.

We observe a consistent improvement across all the evaluated setups when following the proposed approach compared to the evaluated baseline. When considering no TWIST fine-tuning of the \ac{LM}, \method significantly outperforms the k-means alternative. Interestingly, even without applying TWIST to the proposed method it provides competitive performance to TWIST using the k-means tokenizer. 

Another benefit of following \method is it requires significantly fewer tunable parameters compared to TWIST. For example, to obtain the results presented in \Cref{LM-results} \method requires training $~\sim230$M parameters less than TWIST.

\begin{table}[t]
\centering
\begin{tabular}{lccccccc}
\toprule
\textbf{Model} & \multicolumn{4}{c}{\textbf{WER}$\downarrow$} &  \multicolumn{2}{c}{\textbf{ABX}$\downarrow$}\\
   & test & test & dev & dev & \multirow{2}{*}{acr.} & \multirow{2}{*}{wit.}\\
   & clean & other & clean & other &  & \\
 \midrule
k-means\rid{1} & 6.83  & 15.57  & 6.59 & 15.25 & 5.88 & 4.54 \\
\method\rid{4} & 6.08 & 13.67 & 5.67 & 13.51  & 7.73 & 6.28\\
\bottomrule
\end{tabular}
\caption{
LibriSpeech WER (\%) of different tokenization methods. Subsets clean and other refered as "c" and "o"
}
\label{ASR-results}
\end{table}

\begin{figure*}[t!]
  \centering
  \hfill
  \begin{subfigure}[b]{0.15\linewidth}
    \centering
    \includegraphics[width=\linewidth]{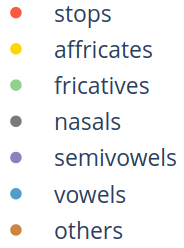}
    \label{fig:amitay_c}
    
  \end{subfigure}
  \hfill
  \begin{subfigure}[t]{0.3\linewidth}
    \centering
    \includegraphics[width=\linewidth]{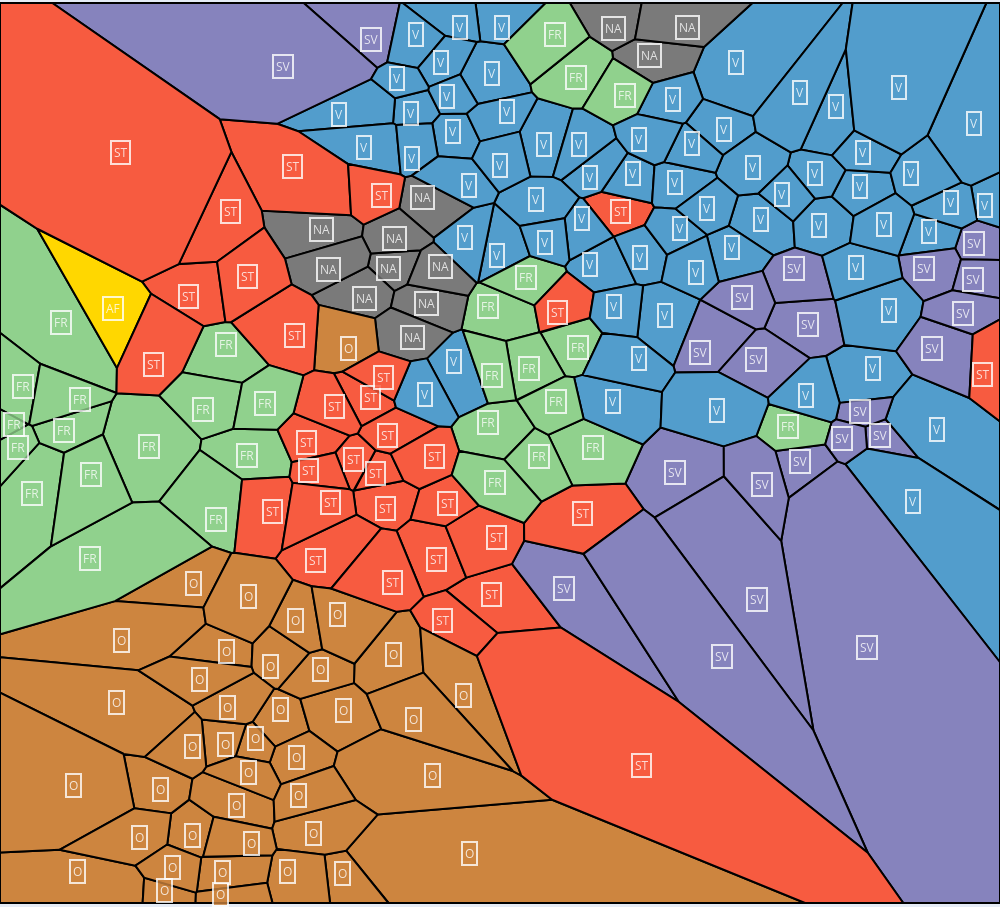}
    \caption{KM}
    \label{fig:amitay_d}
  \end{subfigure}
  \begin{subfigure}[t]{0.3\linewidth}
    \centering
    \includegraphics[width=\linewidth]{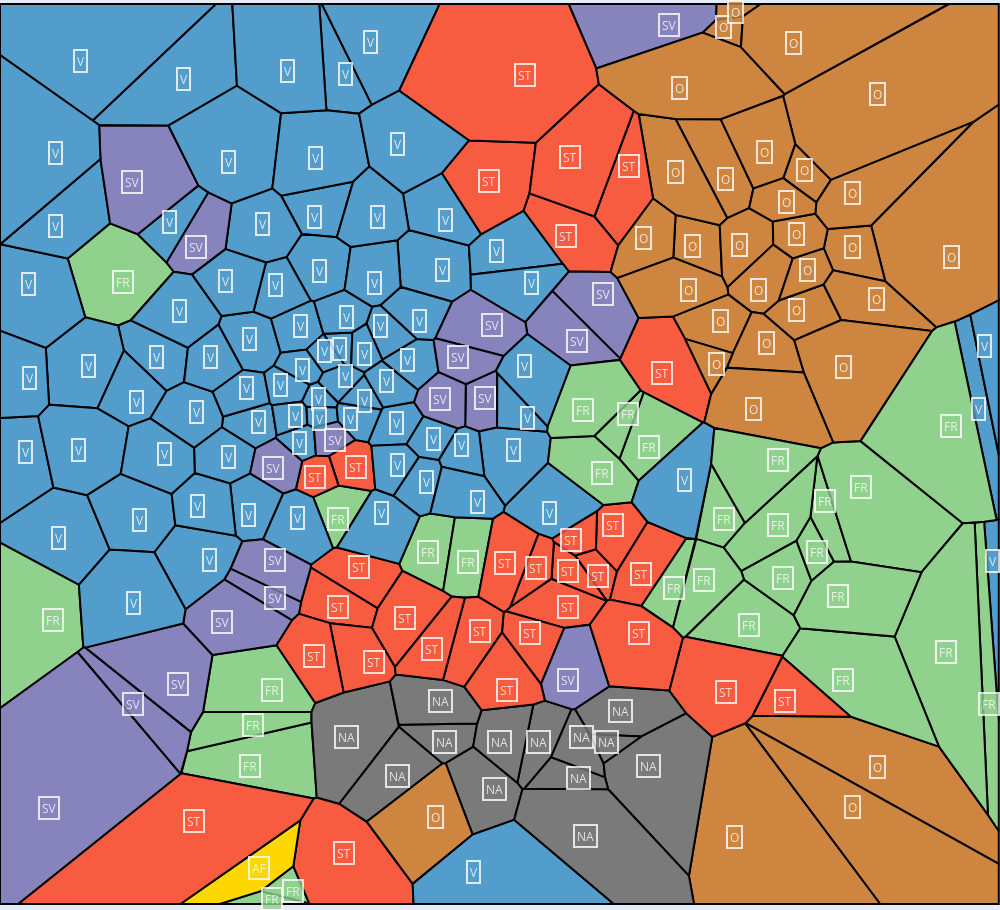}
    \caption{\method}
    \label{fig:amitay_e}
  \end{subfigure}
  
  \caption{Units visualization. Each bounded area represents a single unit out of 200 and is colored by the unit’s phoneme family.}
  \label{fig:amitay}
\end{figure*}

\subsection{Speech Recognition}
\label{subsec:asr}
Next, we evaluate how the proposed tokenization method is able to preserve linguistic content. For that, we train an acoustic model for the task of \ac{ASR} and measured the \ac{WER}. We train a T5-\texttt{base} \cite{2020t5} seq2seq model on LS-$960$. Text tokenization is similar across all methods using T5-\texttt{base} pre-trained tokenizer.

We additionally measure the ABX metric, as proposed by~\cite{dunbar2017zero}. Unlike \ac{WER} which measures the content preservation at the word level, ABX operates at the phoneme level. The ABX task measures the phonetic discriminative abilities of the representation. It involves a pair of words differing by a single phoneme and a reference test word sharing a phoneme with one of the pair. It assesses whether the test phoneme is closer in representation to the correct or incorrect phoneme, expecting a shorter distance to the correct one. The ABX task is conducted in two setups: 'within' and 'across'. 'Within' is evaluated on input data from the same speaker, while 'across' is evaluated on input data from different speakers. Notice, our goal is not to construct the best performing \ac{ASR} model~\cite{chang2023exploration,chang2023exploring}, we are primarily interested in comparing the relative results of \method compared to the k-means alternative.

Results are presented in \Cref{ASR-results}. When considering the \ac{WER} metric, \method shows superior performance than the k-means alternative, however, when considering the ABX metric, the k-means method outperforms \method. Considering the way both methods were constructed, these findings imply that the k-means model better captures phonemic information.

To better analyze this, we follow \cite{sicherman2023analysing} and visualize the tokens clusters as a function of phoneme families measured over the TIMIT corpus which contains full alignment. For each token cluster, we select the majority phoneme, the one that was most present in the cluster. Results are depicted in \Cref{fig:amitay}. Results suggest that both methods are correlative with phonetic information.

\section{Ablation}
\label{sec:abl}
We turn into analyzing the different components composing \method. We start by evaluating the effect of the vocabulary size, we systematically compare different numbers of codebook sizes. Next, we evaluate the effect of the LM tokenizer used and the final \slm. We consider LM of different sizes from the OPT family of models. Lastly, we consider different numbers of layers for encoding, decoding and modality adapting layers.

\subsection{Vocabulary Size \label{subsec:vocab_size}}

Similarly to \cite{hsu2021hubert,lakhotia2021generative,hassid2024textually} we experiment with different codebook sizes to better evaluate the effect of the number of tokens on modelling performance. We measure zero-resource speech metrics (i.e., sWUGGY and sBLIMP) using \slms with the additional decoding layers. Results are reported in \Cref{table:tokens_freq}. Results suggest that $500$ tokens provide the best performance. Interestingly using $1000$ units provides inferior performance to even $100$ units. We suspect this is due to high redundancy in the codebook usage.

\begin{table}[t!]
\centering
\begin{tabular}{cccccc}
\toprule
\textbf{\# Tokens} & \textbf{sWUGGY} & \textbf{sBLIMP} \\
\midrule
100\rid{19}   & 68.55 & 54.82 \\
200\rid{18}   & 70.66 & 54.08 \\
500\rid{6}    & \textbf{71.78} & \textbf{55.25} \\
1000\rid{26}  & 67.77 & 54.64 \\
\bottomrule
\end{tabular}
\caption{\method zero-shot modeling results for different numbers of tokens.
}
\label{table:tokens_freq}
\end{table}

\subsection{Tokenizer-Model Relation}

Next, we evaluate the effect of LM size on the tokenization method and as a \slm afterwords. We consider different model sizes from the OPT model family (i.e., $125$M and $350$M parameters) as either the \slm and the model used for tokenization (denoted by Tokenizer LM). Results are reported in \Cref{table:model-tokenizer}. 

We observe several interesting findings: (i) although conducting the experiments on the same model family, we can say our tokenizers can be used by different \ac{LM}s. (ii) finetuning the $125$M with the tokenizer trained with the $350$M LM achieves better results than the one finetuned with the tokenizer train with the $125$M LM.

\begin{table}[t!]
\centering

\begin{tabular}{llcc}
\toprule
\textbf{Model LM}& \textbf{Tokenizer LM} & \textbf{sWUGGY} & \textbf{sBLIMP} \\
\midrule
OPT-125M\rid{43} & OPT-125M\rid{44} &  70.81    & 55.13 \\
                 & OPT-350M\rid{45} &  \textbf{72.66}    & \textbf{55.27} \\                
\midrule
OPT-350M\rid{6} & OPT-350M\rid{25,42}&  \textbf{74.24} & \textbf{56.80} \\            
\bottomrule
\end{tabular}

\caption{
 We evaluate the effect of the \ac{LM} size used training \method on the zero-shot modeling results finetuning pretrained text \ac{LM}.
}
\label{table:model-tokenizer}
\end{table}

\subsection{Modelling Choices}
\label{subsec:adaptors}

\newpara{Model Architecture.} We analyze the effect of different architectural choices of \method. Specifically, we consider different numbers of layers in the encoder, decoder, and adaptation modules. For the adaptation module, we additionally consider a varying number of layers before and after the LM. In all setups, we report the sWUGGY metric using OPT-$350$ with $500$ tokens. Results presented in \Cref{table:adaptors}. 

Results suggest that except of one setup (second row in \Cref{table:adaptors}), the proposed method is not sensitive to the specific design choice. However, a few interesting insights comes up: (i) using $2$ layers for all modules provides the best overall results; (ii) using more adaptation layers before the LM provides superior performance than using more adaptation layers after the LM; and (iii) using equal number of adaptation layers before and after the LM provides the best performance and stabilizes the results, as long as there is enough capacity in the encoder module. 

\newpara{Reconstruction loss.} Lastly, we discussed the need for the L2 loss \Cref{subsec:training} to prevent collapse during model optimization. Specifically, we proposed to regularize the training process with an L2 loss between the $D(u)$ and $v$. To evaluate different regularization alternatives, we experiment with replacing $v$ with the different layer from $f$ (HuBERT model), specifically layer $10$ or with the acoustic features obtained from the first CNN layers in HuBERT. However, we found both options to provide inferior performance to the currently used one. 

\section{Related Work}
\label{sec:related}

\newpara{Speech Language Models.} Speech language models were first demonstrated by \cite{on_generative}. The authors showed how raw and uncurated speech data can be leveraged into building a \ac{GSLM} system. Next,~\cite{kharitonov2021text} proposed a multi-stream \slm to jointly process ``pseudo-text'' tokens together with quantized prosodic features (i.e., duration and F0). Such a modeling framework opened up a new and promising research direction for processing and modeling spoken data. \cite{polyak2021speech} evaluated the robustness and disentanglement properties of speech-to-tokens models and demonstrated the ability to perform voice conversion as well as a lightweight speech codec. \cite{kreuk2021textless} proposed to cast the task of speech emotion conversion as a translation task, hence translating between one emotion to the other in the discrete space, while \cite{maimon2022speaking} proposed a similar approach for speaking style conversion. \cite{nguyen2022generative} proposed training two \slms jointly to mimic natural spoken dialogues. Recently, \cite{borsos2022audiolm} proposed cascading several LMs, in which one LM operates over semantic speech tokens while the others operate on acoustic tokens. Such modeling framework allows generating natural speech while keeping the identity of the speaker and acoustic conditions unchanged~\cite{wang2023neural, kharitonov2023speak}. \cite{sicherman2023analysing} show that the semantic units obtained from such models highly correlate to phonemes and phoneme states. \cite{lee2021direct, lee-etal-2022-textless, popuri2022enhanced, peng2024mslm} followed a similar modeling mechanism using a different speech tokenizer and proposed a textless approach for speech-to-speech translation. \cite{bapna2021slam, cheng2022mu} and \cite{bapna2022mslam} considered speech as another language in multilingual setups, and showed that involving speech and text as part of training data improves results for speech translation and multilingual text tasks. \cite{Ao2021SpeechT5, chen2023maestro} and \cite{rubenstein2023audiopalm} used the joint training to improve transcriptions tasks as \ac{ASR} and \ac{TTS}. \cite{nachmani2023spoken} proposed augmenting a text \ac{LM} with continuous speech data to improve spoken question-answering tasks. 

\begin{table}[t!]
\centering
\begin{tabular}{ccccc}
\toprule
\textbf{\# Enc} & \textbf{\# Dec} & \# \textbf{LM-AB} & \# \textbf{LM-AA} & \textbf{sWUGGY}  \\
\midrule
\rid{53}2 &2  &  1 & 1 & 68.78  \\
\rid{54} 2 &2  &  1 & 3 & 59.21  \\
\rid{55} 2 &2  &  3 & 1 & 69.47  \\
\rid{6}2 &2  &  2 & 2 & \textbf{71.78}  \\
\rid{59}3 & 1 &  2 & 2 & 70.31  \\
\rid{60}1 & 3 &  2 & 2 & 67.78  \\
\rid{56}1 & 1 &  2 & 2 & 70.39   \\
\bottomrule
\end{tabular}
\caption
{
Model architecture ablation. We compare \method variations with different numbers of trainable layers across different modules. \# Enc and \# Dec stands for the number of encoder and decoder layers in the adaptors respectively. \# LM-AB and \# LM-AA stands for the number of layers in the adaptation layers before and after the the LM respectively. 
}
\label{table:adaptors}
\end{table}

\newpara{Speech and Audio Tokenizers.} The most common approach to tokenize speech for spoken language modeling is the k-means method, which was first proposed by~\cite{on_generative}. Later on, other studies follow a similar modeling paradigm~\cite{borsos2022audiolm, kreuk2021textless, kharitonov2021text, wu2023improving}. These results in a discrete semantic representation of the speech signal. \cite{gat2022augmentation, messica2024nast} pointed out the lack of robustness in current k-means tokenization methods and proposed to match representations from clean and augmented signals. Another alternative approach is to follow the VQ-VAE approach using an RVQ~\cite{defossez2022high, zeghidour2021soundstream}. Due to its objective function, such representation better captures acoustic information rather than semantic information. Recent studies either use the semantic tokens~\cite{on_generative, hassid2024textually}, the acoustic tokens conditioned on global textual descriptions ~\cite{kreuk2022audiogen, copet2024simple, wang2023neural}, or both semantic and acoustic tokens~\cite{borsos2022audiolm}. Recently,~\cite{zhang2023speechtokenizer} proposed to jointly train speech tokenizers to capture both semantic and acoustic tokens. The authors proposed training a discrete auto-encoder using RVQ and optimizing the first RVQ codebook to be similar to the semantic tokens obtained from the k-means model. Notice, that all of the above methods are orthogonal to the proposed method as they can be applied jointly with our method. 

Another relevant related work to ours is the work by \cite{yariv2023audiotoken, yariv2024diverse} that proposes to leverage a pre-train text-to-image and text-to-video models (respectively) to build image and video generation models condition on audio inputs. The authors proposed to augment the textual inputs with the newly learned single audio token. Unlike these works, our method is focused on speech and not general audio, and more importantly, aimed at learning a full vocabulary over time rather than a single audio token.
\section{Discussion}
\label{sec:conclusion}
In this study, we propose \method: a language model aware speech tokenization method. Unlike prior work, which constructs the speech tokens independently from the \slm, \method involves the \slm during the tokenization process. \method leverages both frozen pre-trained contextualized speech encoder and frozen pre-trained text \ac{LM} while introducing a lightweight modality adapter. Results suggest \method provides superior performance to the k-means alternative considering both zero-resource metrics and transcription capabilities. Interestingly, as \method augments a frozen pre-trained text \ac{LM}, it is not only superior in terms of speech modeling but also keeps the same text capabilities of the original \ac{LM} model. 

\newpara{Limitations.} While \method is superior to the k-means alternative and allows to optimize jointly the speech tokenizer and \slms, it requires more computational resources than the standard k-means approach. Additionally, we only presented results for zero-resource speech modeling metrics, more evaluations are needed in the direction of unit-to-speech synthesis. We leave that for future research.  

\newpara{Future work.} We intend to extend this work in two main directions: (i) as mentioned above, evaluating \method also under the unit-to-speech synthesis framework. This will allow us to evaluate the system similarly to the \ac{GSLM} setup as was done by \cite{on_generative}. Additionally, as the demonstrated that a frozen text \ac{LM} can be efficiently augmented with speech tokens, we intend to explore the merger of the two modalities in the form of \ac{ASR} and \ac{TTS}.

\newpara{Acknowledgements.} This research work was supported by ISF grant 2049/22.

\bibliographystyle{IEEEbib}
\bibliography{bib,refs}

\end{document}